\documentclass{article}

\usepackage[square,numbers,compress]{natbib}
\PassOptionsToPackage{square, numbers, compress}{natbib}

\usepackage[preprint]{neurips_2021}

\usepackage[utf8]{inputenc} 
\usepackage[T1]{fontenc}    
\usepackage{hyperref}       
\usepackage{url}            
\usepackage{booktabs}       
\usepackage{amsfonts}       
\usepackage{nicefrac}       
\usepackage{microtype}      
\usepackage{xcolor}         

\usepackage[pdftex]{graphicx}

\usepackage{enumitem}

\usepackage{array}
\newcolumntype{+}{>{\global\let\currentrowstyle\relax}}
\newcolumntype{^}{>{\currentrowstyle}}

\usepackage[bottom]{footmisc}

\title{espiownage: Tracking Transients in Steelpan Drum Strikes Using Surveillance Technology}

\author{%
  Scott H. Hawley\thanks{scott.hawley@belmont.edu} \\
  Department of Chemistry \& Physics\\
  Belmont University\\
  Nashville, TN 37212 \\
  \texttt{scott.hawley@belmont.edu} \\
  \And
  Andrew C. Morrison \\
  Natural Science Department \\
  Joliet Junior College \\
  Joliet, IL 60431 \\
  \AND
  Grant S. Morgan \\
  Department of Chemistry \& Physics\\
  Belmont University\\
  Nashville, TN 37212 \
}

\begin{document}

\maketitle

\begin{abstract}
We present an improvement in the ability to meaningfully track features in high speed videos of Caribbean steelpan drums illuminated by Electronic  Speckle Pattern Interferometry (ESPI). This is achieved through the use of up-to-date computer vision libraries for object
detection and image segmentation as well as a significant effort toward cleaning the dataset previously used to train systems for this application. Besides improvements on previous metric scores by 10\% or more, noteworthy in this project are the introduction of a segmentation-regression map for the entire drum surface yielding interference fringe counts comparable to those obtained via object detection, as well as the accelerated workflow for coordinating the data-cleaning-and-model-training feedback loop for rapid iteration allowing this project to be conducted on a  timescale of only 18 days. 
\end{abstract}

\section{\label{sec:1} Introduction}
\vspace{-0.1cm}
\paragraph{ESPI in Musical Acoustics.}
In the field of musical acoustics, {Electronic Speckle Pattern Interferometry (ESPI)} has shown to be a useful tool. Vibrating plates and membranes in musical instruments such as violins, guitars, drums, and other instruments can be measured and visualized using ESPI.\citep{Moore2018,Bakarezos2019} Small vibrations can be measured using ESPI; time-averaged amplitude measurements are possible. Light and dark fringes, which are lines of constant surface deformation proportionate to the wavelength of the laser light, appear in ESPI photographs. In that they expose the mode forms of vibrating surfaces, these images are analogous to Chladni patterns, however Chladni patterns are typically associated with standing wave patterns. Images of rapid transient phenomena require greater sophistication and their interpretation has to date received little attention in musical acoustics.



\paragraph{Prior Work.}
Recent work by Hawley \& Morrison \citep{hm2021} showed the use of a neural-network-based object detector to track transient phenomena in Caribbean steelpan drums illuminated by ESPI and filmed at 15,037 frames per second\ \citep{morrison_high_2011}. Individual video frames were annotated by crowdsourced human volunteers as part of the ``Steelpan Vibrations Project" (SVP)\ \citep{SVP} in partnership with the Zooniverse.org \citep{borne_zooniverse_2011}, but human annotation progress was slow. Thus the object detector was trained on existing annotations and used to predict annotations for the remainder of the videos, yielding preliminary physics results. Their object detector was a custom-written scheme based on YOLO9000\ \citep{YOLO9000}, predicting elliptical antinode regions and a regression-based count of the interference ``rings'' appearing in each antinode.  The study was limited by low ``regression accuracy'' scores for the real data  compared to synthetic or ``fake'' data. This was attributed to having noisy or ``unclean'' annotations, and the `preliminary physics'' conclusions were reported without full confidence.  Further hindering the project were software engineering limitations such as the use of libraries making it difficult to maintain, and it was unable to perform Transfer Learning \citep{Wang2020Pay}, requiring a time-consuming training from scratch. 

\paragraph{This Study.}
Upon learning of the extension of the deadline for this workshop\ \citep{cranmertweet}, we embarked on an attempt to improve upon prior work by taking advantage of newer models and software environments as well as best practices for data-cleaning that involved rapid iteration between model training and human (re-)annotations.  Key results and features of this effort are as follows:

\begin{enumerate}[noitemsep,nolistsep]
  \item Better scores for both ring-count accuracy and COCO mAP scores than prior work. These were obtained by separating the task of detecting antinodes from the counting of rings in cropped sub-images.  Bounding boxes (rather than ellipses) were detected around antinodes. These boxes were then used to crop the images. Then the ring-counts were obtained for the cropped (sub-)images.
   \item The introduction of a ``segmentation regression'' mapping method for ring counting which produces results in close agreement with the values from the crop-and-count method.
  \item The release of a new dataset for ``real data'' of annotations of ESPI images of steelpan drums.
   \item The exploration -- in Supplemental Materials due to space limitations -- of the generalization of models trained on steelpan drum images to other musical instruments, including those with non-elliptical antinode regions.
  \item The {\it timescale} of this project and its methodology for rapid data cleaning by establishing a feedback loop between model predictions and the graphical data-editor software, prioritizing ``top loss'' examples so that human data-cleaners' efforts could be directed efficiently. 
  \item Both the timescale and improved metrics necessitated building on up-to-date, well-maintained ML development libraries such as fast.ai\ \citep{fastai} and IceVision\ \citep{icevision2020} ({\it i.e.,} over custom-written object dectection code).
  This allowed the quick integration of capabilities such as transfer learning, newer optimizers\ \cite{Ranger}, and run logging\ \cite{wandb}. It is hard to overstate the utility of the nbdev\citep{nbdev} development system for ``literate programming,'' allowing code, documentation, and examples all to be written {\it as one} as Jupyter notebooks\ \citep{jupyter} and posted for immediate use by collaborators via Google Colaboratory.\footnote{For reproducibility, anonymized documentation of results in the form of executable Colab notebooks are hosted at \url{https://drscotthawley.github.io/espiownage/}, plus Supplemental Materials such as movies.
  }
\end{enumerate}

\begin{figure}[hb!]
\vspace{-0.4cm}
  \centering
      \includegraphics[width=0.42\linewidth]{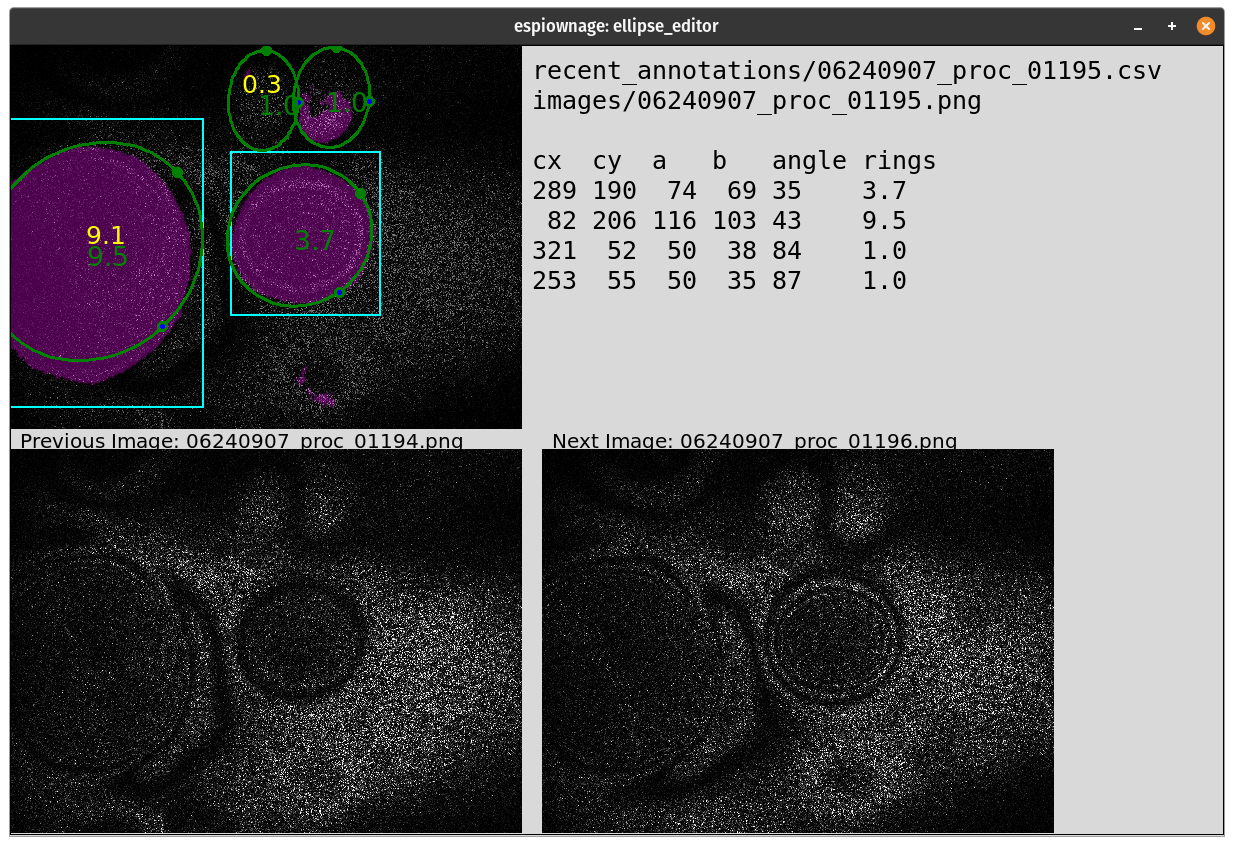} 
    \includegraphics[trim={0 0.2cm 0 0},clip,width=0.57\linewidth]{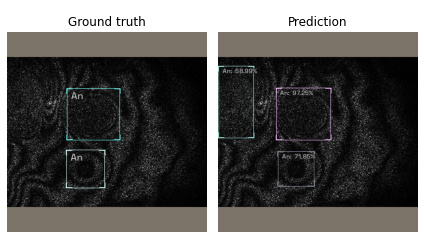} 
\caption{
Left: Screenshot of our enhanced ``ellipse editor`` tool, which builds on code released in prior work\ \citep{hm2021}. Besides the prior ability to graphically edit annotations of boundaries and ring counts elliptical antinode regions, this newer version of the software displays predictions from the neural network models' predictions of bounding boxes, ring counts, and segmentation regression maps.
Middle: Bounding box detection of antinode regions via IceVision\ \cite{icevision2020} using their tuned RetinaNet\ \citep{retinanet} model.  We also tried detecting individual rings as objects but there were too many false negatives, whereas the model was almost always able to detect entire antinodes, including those that annotators missed. The cropped regions became inputs to the ring-counting code, a binary classifier adapted for regression by extending the vertical range of the final sigmoid to the range of our outputs ({\it i.e.,} we stay within the linear regime of the sigmoid). Note that the antinodes basically circular, becoming more so when cropped and re-shaped as square images, which then allows for arbitrary rotations in addition to other standard image data augmentation methods.}
\label{fig:boxes_and_crops}
\end{figure}

\section{Data Cleaning Workflow}
\vspace{-0.1cm}
Hawley \& Morrison\ \citep{hm2021} offered evidence suggesting that inconsistent annotations in their dataset were primarily responsible for poor performance on their metric of ``ring count accuracy,'' a classification-like score counting ring counts withing $\pm 0.5$ rings of each other as a match. We sought to achieve better metric scores via more intensive data-cleaning, as well as to explore whether that metric of $\pm 0.5$ was perhaps too stringent to properly reflect the model's performance and ``believability'' as a tool for discovering the physical dynamics of steelpan drum transients.  Figure \ref{fig:boxes_and_crops} shows a screenshot of the interactive graphical tool used to clean the dataset. 
Though we were granted access to the hand-edited "SPNet" dataset of \citep{hm2021}, we chose to start fresh from the aggregated annotations of 15 or more volunteers on Zooniverse\ \citep{borne_zooniverse_2011}. This turned out to be at least as noisy as the SPNet dataset, and we too needed data cleaning. For this paper, we refer to the original state as the "Pre-cleaned" dataset, and show its differences from both our final "Real" dataset and the SPNet Real dataset. In order to improve the efficiency of the data-cleaning effort, we modified the ellipse editor in order to show model predictions as users cleaned the data, ordering the images seen according to decreasing loss values.
See the caption for Figure \ref{fig:boxes_and_crops} for additional details on data cleaning.


\begin{table}[hb!]
\vspace{0.2cm}
  \caption{Scores for bounding box prediction COCO mAP\ \citep{coco} and ring counts from the cropped images. The "$\pm X$" columns are accuracy scores for predictions within $X$ rings of the target. Uncertainties are $\pm 1$  in the last digit or less, except the Pre-cleaned dataset for which they are $\pm 2$. Low scores are better for Mean Absolute Error (MAE) whereas high scores are better for other columns.}
    \centering
    \begin{tabular}{llllllll}
    \toprule
        Dataset  &  mAP & MAE & $\pm$0.5 & $\pm$0.7 & $\pm$1 & $\pm$1.5 & $\pm$2  \\ 
        \midrule
        SPNet Real & 0.68 & 0.85 & 0.43 & 0.55 & 0.70 & 0.83 & 0.91  \\ 
        Pre-cleaned & 0.66 & 0.96 & 0.41 & 0.52 & 0.66 & 0.79 & 0.87  \\ 
        \bf Our Real   & \bf 0.68 &\bf  0.71 &\bf  0.53 &\bf  0.66 &\bf  0.79 &\bf  0.89 &\bf  0.94 \\ 
        SPNet CycleGAN   & 0.73 & 0.18 & 0.93 & 0.96 & 1.0 & 1.0 & 1.0  \\ 
        Our Fake   & 0.865 & 0.21 & 0.93 & 0.97 & 0.99 & 1.0 & 1.0   \\ 
        \bottomrule
    \end{tabular}
\label{tab:boxes_and_crops}
\end{table}

\begin{figure}[h!]
\centering
  \includegraphics[width=.9\linewidth]{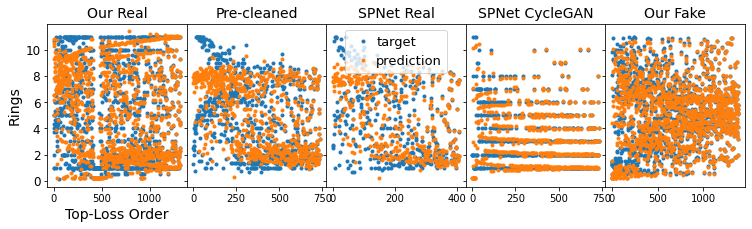} \\  
  \vspace{-0.1cm}
   \includegraphics[width=.9\linewidth]{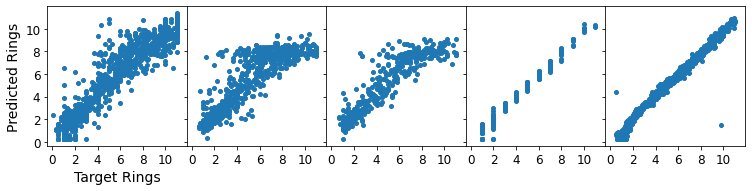} 
\caption{Ring counts obtained from cropped images. Top: predicted and target values, arranged in order of worst agreement to best, for the different datasets. The prevelance of values at the max (11) and min ($\sim$1) are reflections of the data-annotation policy of the SVP\ \citep{SVP}. Note how the CycleGAN dataset from \citep{hm2021}, despite its visual similarity to the real images, only contained integer ring values. Bottom: Plots of predicted rings vs. target rings, showing that our data-cleaning effort (``real,'' left column) resulted in closer agreement and less compression of the dynamic range that other datasets.}
\label{fig:ring_counts}
\end{figure}

\begin{table}[]
\vspace{0.6cm}
  \caption{Table of scores for segmentation regression maps for the case of mask quantization with a bin size of 0.7, which was the minimum resolution at which the model was able to train effectively. Uncertainties are the same as in Table \ref{tab:boxes_and_crops}. For CycleGAN data, training stagnated immediately. The scores here are lower than in Table 1, because they are integrated over the entire image. See Figure 4 (left pane) for a comparison showing agreement between the different methods.}
  
    \centering
    \begin{tabular}{llllllll}
    \toprule
        Dataset  &  MAE & $\pm$0.5 & $\pm$0.7 & $\pm$1 & $\pm$1.5 & $\pm$2  \\ 
        \midrule
        SPNet Real & 0.61 & 0.19 & 0.26 & 0.36 & 0.50 & 0.62  \\ 
        Pre-cleaned & 0.74 & 0.15 & 0.21 & 0.29 & 0.41 & 0.52  \\ 
        \bf Our Real   &\bf  0.75 &\bf  0.19 &\bf  0.25 &\bf  0.37 &\bf  0.49 &\bf  0.62 \\
        SPNet CycleGAN   & 0.20 & 0.00 & 0.00 & 0.00 & 1.00 & 1.00  \\ 
        Our Fake   & 0.19 & 0.52 & 0.66 & 0.80 & 0.90 & 0.94   \\ 
        \bottomrule
    \end{tabular}
\label{tab:segreg}
\end{table}

\begin{figure}[hb!]
\begin{tabular}{lr}
  \centering
   \includegraphics[width=0.5\linewidth]{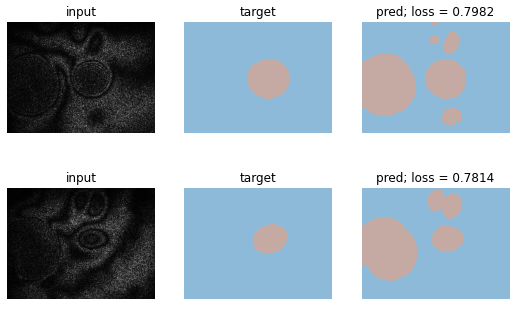} &
    \includegraphics[width=0.42\linewidth]{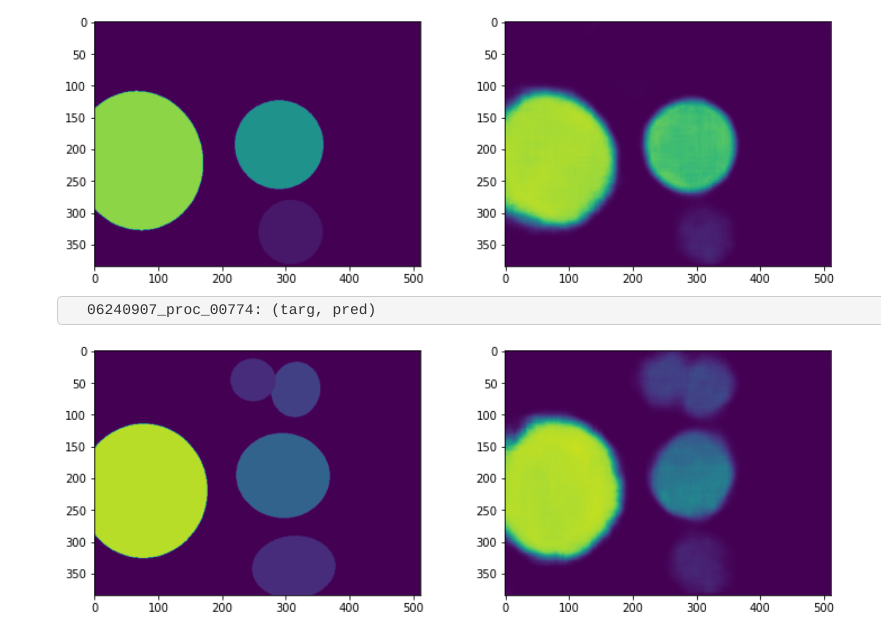}
    \end{tabular}
\vspace{-0.1cm}
  \caption{Segmentation examples. Left: All antinodes as one class, used for inclusion in the ellipse editor for the data-cleaning workflow. Right: Segmentation-regression maps for target and predicted ring counts. In each case there is one output "class" but for regression this is a floating point value and the scale of the final sigmoid activation is scaled to keep the output within the linear regime.
}
\label{fig:segreg_samples}
\end{figure}

\begin{figure}[h!]
  \centering
     \includegraphics[width=0.31\linewidth]{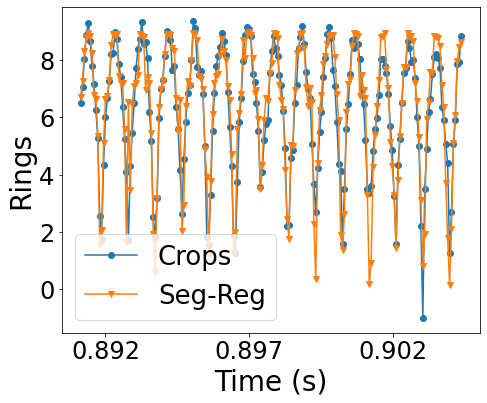} 
     \includegraphics[width=0.34\linewidth]{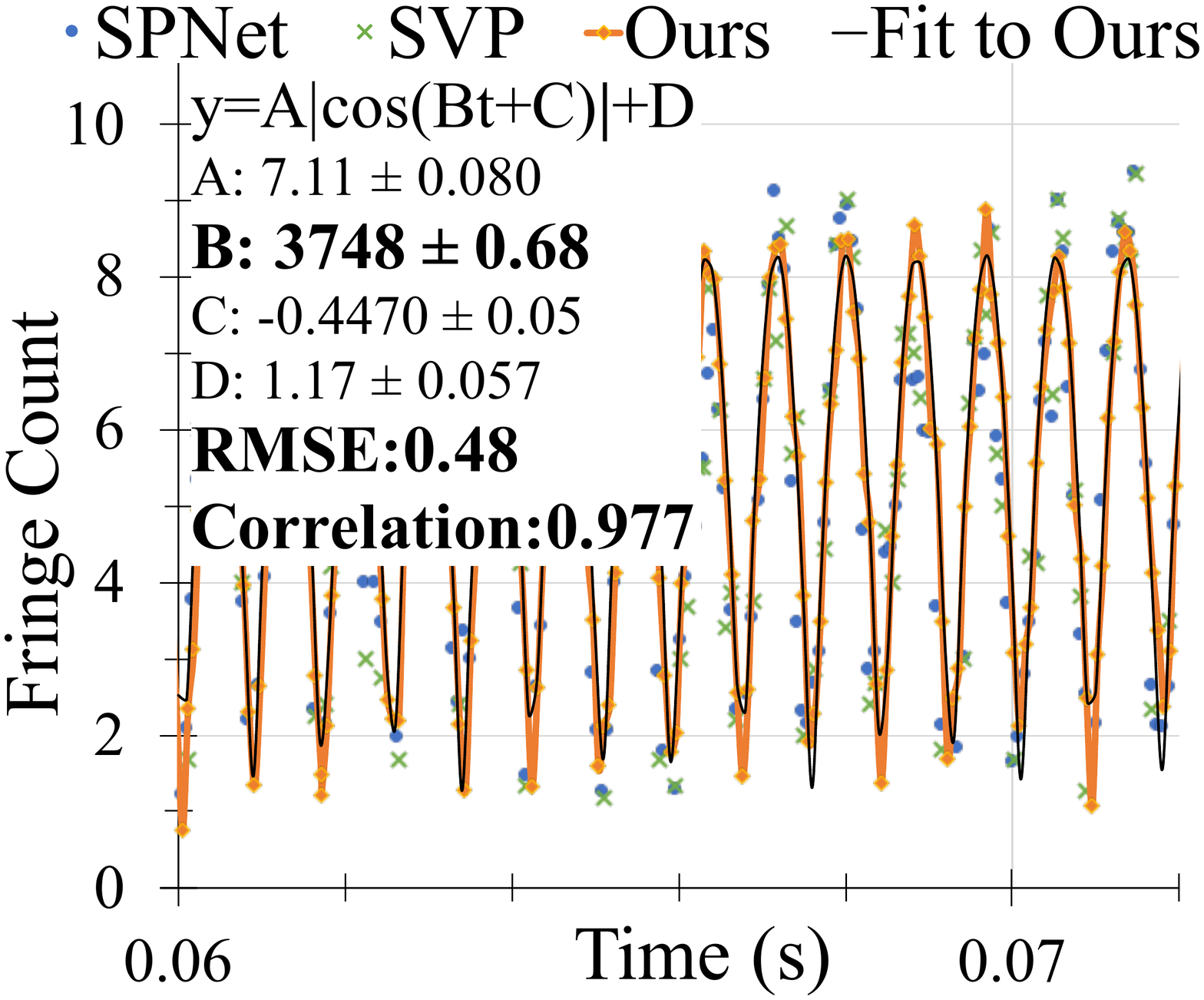}
     \includegraphics[width=0.28\linewidth]{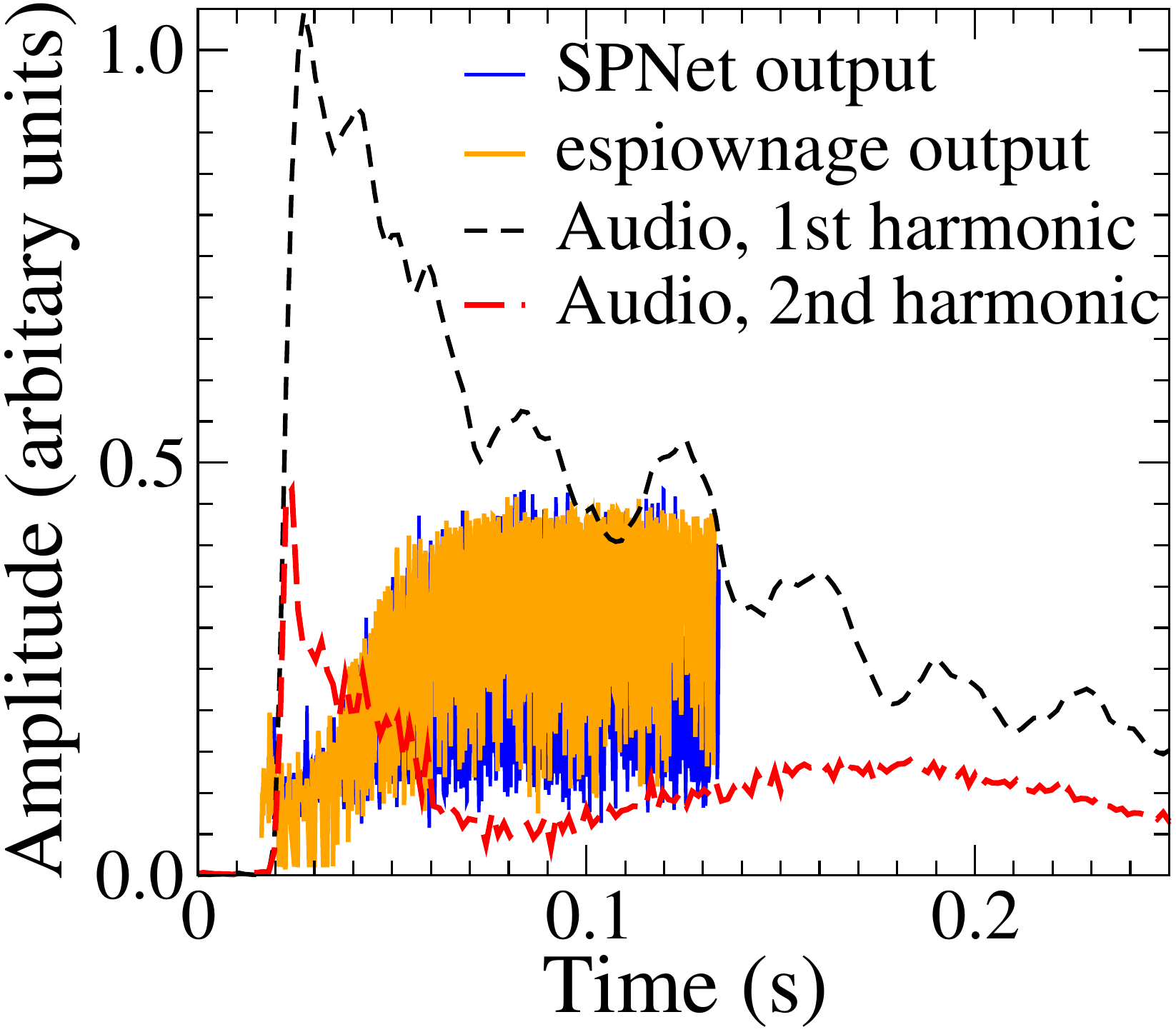}
\vspace{-0.2cm}
\caption{Left: Close agreement between the ring counts (as a function of time) from the bounding-box-and-crop method and the segmentation regression map. Middle: Our version of Figure 6 from \cite{hm2021} yet with lower uncertainties and higher correlation coefficient. Right: Our replication of a key ``preliminary physics'' result in a panel from Figure 7 of \citep{hm2021}, confirming the dissimilarity in rise times between the amplitude heard in audio recordings  (Dot-dashed/black and dashed/red lines) and the amplitude observed in ESPI video.}
\label{fig:curve_fits}
\end{figure}

 \begin{figure}[]
 \vspace{0.2cm}
  \centering
     \includegraphics[width=0.47\linewidth]{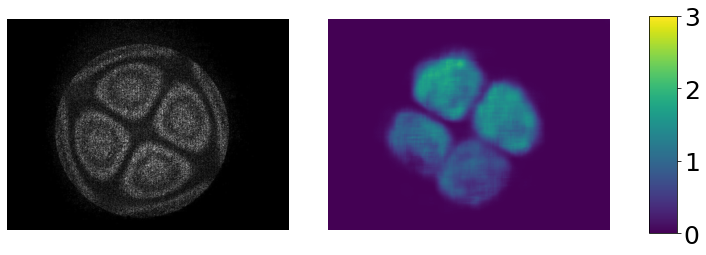}
     \hspace{0.6cm}
     \includegraphics[width=0.47\linewidth]{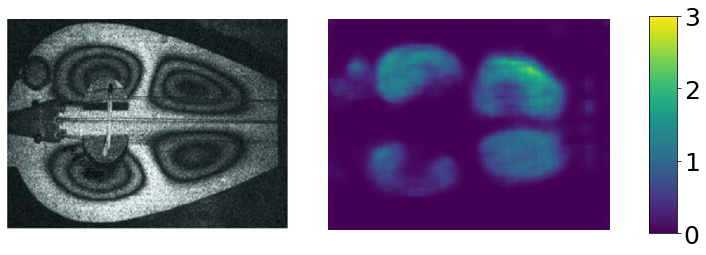}
\caption{Segmentation-regression maps for other instruments, highlighting the generalization performance of our model. 
These maps were obtained via inference using the model trained only on our steelpan dataset. 
Although the model was only trained on elliptical antinode regions, we see that it is able to trace the contours and count rings for the roughly triangular antinode regions for the drum image on the left \citep{moore_ajp} and the bean-shaped antinodes of the 17th-century lyra on the right \citep{Bakarezos2019}.}
\label{fig:other_instruments}
\end{figure}

\section{Results}
We show example images of predicted bounding boxes and sample cropped rings in Figure \ref{fig:boxes_and_crops}, followed by Figure \ref{fig:ring_counts} which shows predicted ring counts as compared to their target values.  The COCO mAP \cite{coco} object detection scores for the bounding boxes and the ``regression accuracy'' scores for various ring count tolerances are provide in Table \ref{tab:boxes_and_crops}.  
Examples of segmentation of images are shown in Figure \ref{fig:segreg_samples}; some of these were true image segmentation ({\it i.e.,} classification) which detected the presence of antinode shapes as a single class. More interesting are the regression maps (right images in the figure) which predict actual ring counts.  
Because the segmentation U-Net method we used in fastai\ \citep{fastai} required integer-valued pixels, we quantized the ring counts at the minimum-trainable resolution of 0.7 rings. Below that, the model would not train.  Figure \ref{fig:curve_fits} demonstrates the efficacy of our methods, and their improvements on prior predictions. Noteworthy is our confirmation of ``preliminary physics'' results in \citep{hm2021}, namely that the non-intuitive differences in timescales between oscillations observed visually and those heard in audio recordings are in fact ``real physics.''

\section{Conclusions}
The methods used in this work improve upon and confirm prior work. A limitation is the scope of application to steelpan drums, though in Supplemental Materials we offer preliminary extensions to other instruments. The algorithms used in this paper are key components of surveillance technology with societal implications when turned on people, however here we have used it to advance the study of musical acoustics.  While all our runs are possible via our provided Colab notebooks, training was performed on two personal workstations with a total of 4 NVIDIA GPUs (RTX 3080, 2080Ti, and two Titan X's). We estimate the total energy expenditure to be $\sim$15 kWh.

\begin{ack}
    The authors thank Zach Mueller for his assistance with {\texttt fastai} and {\texttt nbdev}-based development, and to Farid Hassainia for his help with {\texttt IceVision}.  
     This publication uses data generated via the Zooniverse.org platform, development of which is funded by generous support, including a Global Impact Award from Google, and by a grant from the Alfred P. Sloan Foundation.
\end{ack}

{
\small
\bibliography{steelpan_neurips_2021}
}

\appendix

\section{Appendix}

For additional implementation details, code and documentation, see Supplemental Materials at \url{https://drscotthawley.github.io/espiownage}

\end{document}